# Pediatric Pancreas Segmentation from MRI Scans with Deep Learning

Elif Keles, MD, PhD[1], Merve Yazol, MD[2], Gorkem Durak, MD[1], Ziliang Hong, MS[1], Halil Ertugrul Aktas[1], Zheyuan Zhang, BS[1], Linkai Peng, MS[1], Onkar Susladkar, BS[1], Necati Guzelyel[1], Oznur Leman Boyunaga, MD[2], Cemal Yazici, MD, MS[3], Mark Lowe, MD, PhD[4], Aliye Uc, MD[5], Ulas Bagci, PhD[1]

[1]Department of Radiology, Northwestern University, IL, USA;

[2]Department of Radiology, Gazi University Faculty of Medicine, Ankara, Türkiye;

[3] Division of Gastroenterology and Hepatology, University of Illinois at Chicago, Chicago, IL, USA;

[4] Division of Pediatrics, Gastroenterology, Hepatology and Nutrition, Washington University School of Medicine in St. Louis, MO, USA

[5]Division of Pediatric Gastroenterology, Hepatology, Pancreatology and Nutrition, University of Iowa, Carver College of Medicine, Iowa, IA, USA.

**Corresponding Author:**

Elif Keles, MD- Ph.D.,

Email: elif.keles@northwestern.edu

Phone: 2245035182

737 N. Michigan Avenue, Suite 1600,

Department of Radiology, Feinberg School of Medicine,

Chicago, IL 60611, USA

**Abstract**

**Objective:** Our study aimed to evaluate and validate *PanSegNet*, a deep learning (DL) algorithm for pediatric pancreas segmentation on MRI in children with acute pancreatitis (AP), chronic pancreatitis (CP), and healthy controls.

**Methods:** With IRB approval, we retrospectively collected 84 MRI scans (1.5T/3T Siemens Aera/Verio) from children aged 2–19 years at Gazi University (2015–2024). The dataset includes healthy children as well as patients diagnosed with AP or CP based on clinical criteria. Pediatric and general radiologists manually segmented the pancreas, then confirmed by a senior pediatric radiologist. *PanSegNet*-generated segmentations were assessed using Dice Similarity Coefficient (DSC) and 95th percentile Hausdorff distance (HD95). Cohen's kappa measured observer agreement.

**Results:** Pancreas MRI T2W scans were obtained from 42 children with AP/CP (mean age: 11.73 ± 3.9 years) and 42 healthy children (mean age: 11.19 ± 4.88 years). *PanSegNet* achieved DSC scores of 88% (controls), 81% (AP), and 80% (CP), with HD95 values of 3.98 mm (controls), 9.85 mm (AP), and 15.67 mm (CP). Inter-observer kappa was 0.86 (controls), 0.82 (pancreatitis), and intra-observer agreement reached 0.88 and 0.81. Strong agreement was observed between automated and manual volumes ($R^2$ = 0.85 in controls, 0.77 in diseased), demonstrating clinical reliability.

**Conclusion:** *PanSegNet* represents the first validated deep learning solution for pancreatic MRI segmentation, achieving expert-level performance across healthy and diseased states. This tool, algorithm, along with our annotated dataset, are freely available on GitHub and OSF, advancing accessible, radiation-free pediatric pancreatic imaging and fostering collaborative research in this underserved domain.

1. INTRODUCTION

**Pancreatitis in Childhood: A Growing Public Health Concern.** Pancreatitis is an emerging inflammatory pancreatic disorder of childhood (1, 2). The incidence of pediatric AP has increased over the last ~20 years, affecting an estimated ~ 13 cases per 100,000 children per year (1, 3). Acute pancreatitis (AP) is characterized by sudden inflammation of the pancreas that may resolve spontaneously without structural and functional alterations. However, some children experience recurrent episodes of AP known as acute recurrent pancreatitis (ARP), and some progress to chronic pancreatitis (CP), characterized by irreversible structural and functional changes in their pancreas (1, 3, 4).

**Clinical Challenges and Diagnostic/Imaging Considerations**. The diagnosis of pancreatitis in childhood is primarily clinical, relying on a combination of history, physical examination, laboratory testing, and imaging findings consistent with the INSPPIRE criteria (4). Imaging plays a critical role in diagnosing pediatric CP and monitoring its complications. However, pediatric-specific literature on pancreatic imaging is scarce, with a significant portion of our knowledge and recommendations derived from the existing adult literature (5, 6). Moreover, current pediatric guidelines are grounded in expert opinions rather than evidence-based findings (5, 6). Therefore, it is essential to establish a framework derived from evidence-based imaging in pediatric pancreatitis.

**Deep learning (DL) for pancreas segmentation.** The increasing reliance on deep learning (DL) techniques has revolutionized the field of medical image segmentation (7, 8). Many studies showed that any lesion or pathology detection and disease diagnosis in various medical disciplines could be enhanced by DL (7, 9, 10). DL algorithms demonstrate the ability to recognize image patterns, including those unseen or not effectively utilized in daily routines, enabling them to detect and diagnose various diseases. DL algorithms leverage biomedical images with predictive/ diagnostic and prognostic characteristics while incorporating clinical data. Those DL algorithms deploy different algorithms, including CNN, Transformers, Capsule Network, and Physics-driven systems (11). We have developed new deep-learning algorithms to enhance imaging-based analysis in several key areas. Our algorithms can segment the adult pancreas using CT and MRI scans (12). Additionally, our models segment pancreas in different phases of adult AP in CT (13) and detect peripancreatic edema in adult AP from CT scans (14) and classify pancreatic IPMN (Intraductal Papillary Mucinous Neoplasm) by integrating DL with radiomics techniques (9, 10, 12, 15, 16). Pediatric pancreas segmentation is challenging for several reasons, with most studies conducted on CTs, not MRI (17). First, the pancreas is a relatively small organ, particularly in children(18), and its shape can vary significantly between individuals, making it challenging to identify and segment accurately (10, 19). Second, the pancreas often has low contrast with surrounding tissues, making it difficult to distinguish boundaries, especially on MRIs (10) . Motion artifacts from breathing can blur the organ boundaries as well. Other pathologies or functional abnormalities can further increase the difficulty of identifying pancreas boundaries (10, 20). In addition to these unique challenges, DL has technical challenges warranting a successful segmentation algorithm for pediatric MRIs. For example, training DL models would typically require large, labeled datasets, which can be time-consuming and expensive to create, especially for pediatric cases (21-26). Their performance would be highly dependent on the quality and diversity of the training data (26-28). One of the main premises behind DL segmentation algorithms in other fields was that there would be availability of various imaging data, their segmentation labels, and source code (25). Unfortunately, this is not the case for pediatric pancreas segmentation. Published studies are not only mainly evaluated with CT scans, but they are also in 2D, not appreciating the volumetric nature of the organ.

**Overall goal:** In this study, we introduce a novel application of DL algorithms to achieve precise segmentation of the pediatric pancreas in 3D (volumetric) MRI images, including healthy subjects and those with AP or CP. Our work addresses critical gaps in pediatric imaging and represents several **firsts** in the field of radiology and artificial intelligence:

- For the first time, we successfully segmented the pediatric pancreas in 3D MRI images, marking a significant milestone in pediatric imaging. Our recently developed *PanSegNet* model is adapted for pediatric MRI, offering cross-platform compatibility and high accuracy in pancreas segmentation.
- We also curated and shared the first multi-disease pediatric pancreas MRI dataset with detailed annotations for T2-weighted imaging, consisting of a diverse population of children aged 2-to-19 years. This dataset represents a unique source for pediatric radiology, facilitating research at the intersection of pediatric radiology, gastroenterology, endocrinology, and other related specialties.
- Our study not only forges the use of AI in pediatric pancreas imaging but also lays the groundwork for future evidence-based guidelines in pediatric pancreatitis. By integrating advanced AI techniques into pediatric radiology, this work is poised to significantly impact clinical practice, enabling more precise diagnosis and personalized care for children with pancreatic diseases. With its multidisciplinary relevance, this study is a crucial step forward in enhancing pediatric imaging by applying cutting-edge technology.

- We provide user-interface-based software, source code, and MRI data with corresponding ground truth labels to ensure reproducibility and transparency, catalyzing future advancements in this field.

## 2. METHODS

### 2.1 Dataset

#### 2.1.1 Data collection and inclusion/exclusion criteria

With IRB approval, we retrospectively collected 84 MRI scans (either 1.5T or 3T from Siemens Aera and Verio Scanners) from healthy children who underwent MRI due to the detected hypoechoic liver lesion on abdominal sonography. Additionally, children with AP or CP were diagnosed according to the established clinical criteria between 2015 to 2024 (4). AP was diagnosed when a patient presented with at least two of the following three criteria: abdominal pain, elevated serum amylase, and/or lipase levels at least three times higher than the upper normal limit, and imaging findings were consistent with AP. The CP was diagnosed with imaging findings suggesting CP, and at least one of the following is present: abdominal pain, exocrine pancreatic insufficiency (EPI), or endocrine insufficiency (1, 4). The imaging protocol varied according to the device. It included the T2-weighted HASTE axial with respiratory triggering, T2-weighted (T2W) HASTE axial with respiratory triggering fat-suppressed sequence, TSE T2W axial, and TSE T2W axial with fat-suppressed sequences. Imaging parameters, including slice thickness (3-6 mm), matrix size (320x168 - 384x198), and percentage field of view (66.25-100), were recorded for each patient, and raw images were utilized by the DL segmentation model. All imaging data were anonymized, and biologic sex, age at imaging, and clinical diagnosis were collected. The collected dataset includes data from 42 healthy children and 42 children who were diagnosed with AP or CP. Healthy children's data included 37 T2-weighted (T2W) and 5 T2 fat-suppressed images. Diseased (AP and CP) children's data included 32 T2W and 10 fat-suppressed images. T1W can be used for segmentation purposes, too; however, we have designated T2W images as the baseline sequence because of the better appearance of the pancreatic duct, peripancreatic fluid, and parenchyma (29, 30). **Table 1** presents a summary of biological sex, age, and image acquisition information for the pediatric population whose images we used in our analysis.

**Table 1:** Patients and image acquisition information of the pediatric population used in our experiments.

|  | **Pancreatitis (n)** | **Healthy (n)** |
|---|---|---|
| **Patients (n)** | 42<br><br>Acute pancreatitis: 23 (54,76%)<br><br>Chronic pancreatitis: 19 (45.24%) | 42 |
| **Biologic Sex (F/M)** | 22/20 | 17/25 |
| **Age (mean ± SD) (min-max) (years)** | 11.73 ± 3.9 (3-19) | 11.19 ± 4.88 (2-18) |
| **1.5T** | 24 | 27 |
| **3T** | 18 | 15 |

2.1.2 Curation and Labeling

Two pediatric radiologists and a pediatrician with 13 years of experience curated the T2W sequences from the dataset, selecting images from healthy children and those with pancreatitis. These T2W sequences were converted to the Neuroimaging Informatics Technology Initiative (NIfTI) format to ensure compatibility with our segmentation workflow. One pediatric radiologist with 13 years' experience and one general radiologist with 7 years' experience manually segmented the pancreas on T2W MRI scans using ITK-SNAP imaging visualization tool version 3.8.0 (31). A senior pediatric radiologist with 30 years of experience double-checked the annotations for quality and consistency. Manual segmentations (after consensus and confirmation by senior pediatric radiologist) were used as the reference standards. To assess the robustness of our manual segmentation protocol, we evaluated both intra-observer and inter-observer variations and Cohen's kappa score. Inter-observer analysis was assessed using the different annotations provided by both pediatric and general radiologists for the entire dataset. To determine intra-observer agreement, the same radiologists reassessed 30 randomly selected MRI scans from each group a second time after a four-week wash-out period. These evaluations helped ensure the reliability and reproducibility of the segmentation process, minimizing observer-related bias in the final annotations. The NIFTI images were then uploaded to the *PanSegNet* software, where the algorithm generated initial segmentation masks **(Figure 1)**.

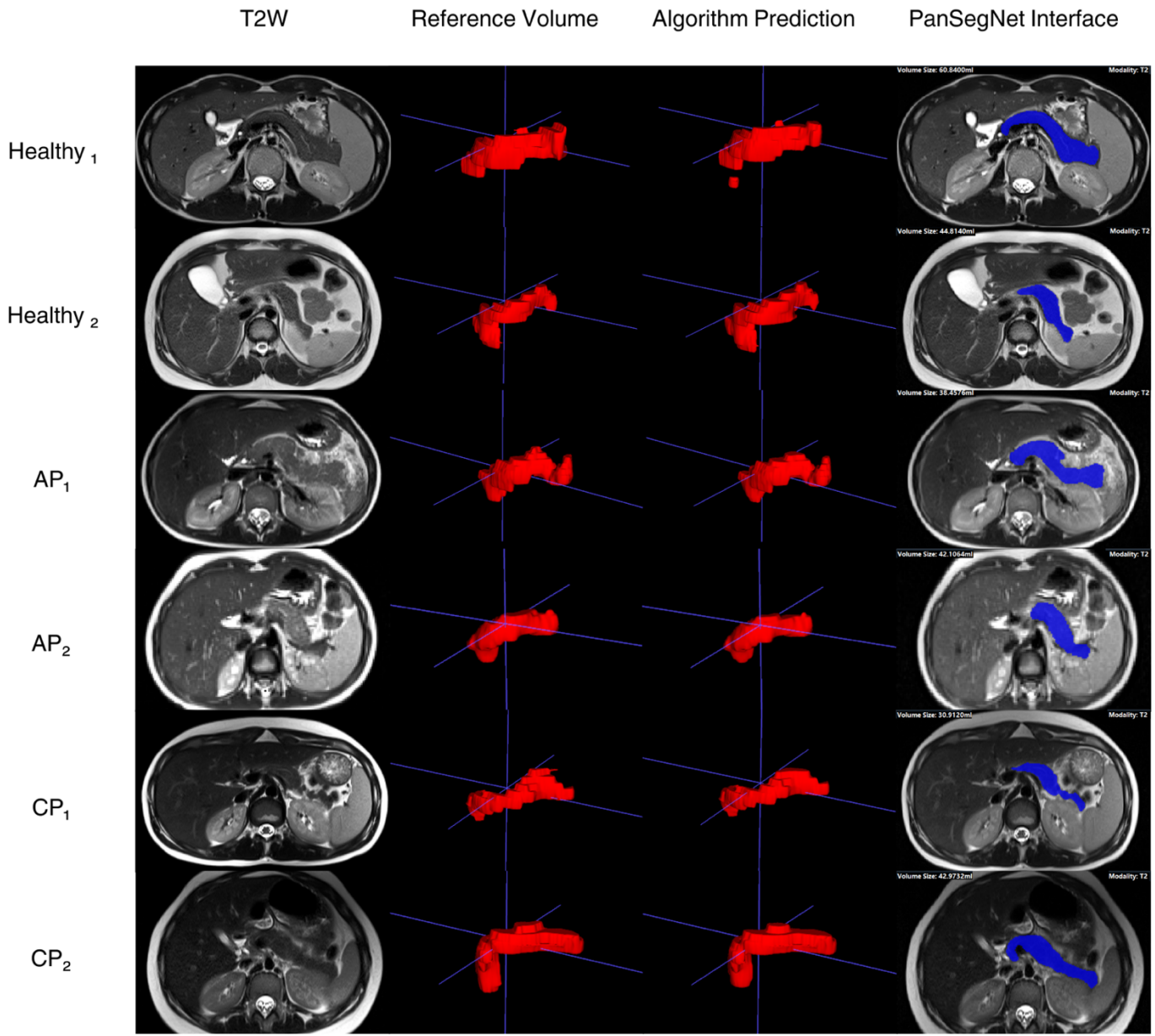

**Figure 1**: Segmentation of the pancreas using *PanSegNet* across healthy controls and patients with acute pancreatitis (AP) or chronic pancreatitis (CP). Columns: (1) T2-weighted MRI images; (2) reference volumes of manual segmentation; (3) predicted segmentation using the *PanSegNet* algorithm; (4) *PanSegNet* segmentation in the software interface. Rows: (1,2) healthy children group; (3,4) children with AP; (5,6) children with CP. Volume rendering was used without smoothing.

2.1.3    Evaluation metrics

In this study, we employ several standard evaluation metrics designed for segmentation, including the **Dice similarity coefficient (DSC)**, **Hausdorff Distance 95th percentile (HD95)**, **Jaccard Index**, Precision, Recall and **the volume prediction error** for the pancreas (7, 32). The DSC assesses the degree of overlap between the predicted segmentation and the ground truths (manual drawings by experts), serving as a key measure of segmentation accuracy; a higher DSC indicates greater overlap and, therefore, higher accuracy (7). The HD95 evaluates the maximum distance between the boundary points of the predicted and actual segmentations, offering insight into boundary precision by highlighting the worst-case error (32). The **Jaccard index** (intersection-over-union) provides a stricter measure of segmentation overlap than DSC, penalizing both false positives and false negatives more heavily, making it particularly useful for assessing segmentation quality in complex pathological presentations (32). **Precision** measures the proportion of predicted pancreatic voxels that are truly pancreas, indicating how well the model avoids false positives—a critical factor in preventing over-segmentation. **Recall**, in contrast, reflects the proportion of actual pancreatic tissue correctly identified by the model, highlighting its ability to fully capture the organ, especially in cases of inflammation or atrophy. These metrics are particularly useful in assessing how well the model captures the fine structural details of the pancreas. Given the clinical significance of pancreas volume in diagnosis and disease progression monitoring, we also calculate the volume prediction error for pancreas segmentation (5, 29, 33, 34). We assessed agreement between automated and manual volume measurements using simple linear regression, referencing manual volumes as the reference standard. The $R^2$ was calculated to quantify how much of the variance in manual volumes could be explained by the automated segmentations. Analyses were performed separately for healthy and diseased groups to evaluate consistency across clinical conditions.

2.1.4   Dataset sharing with the public

Our dataset, called **PedPanSegData,** was fully anonymized and annotated for pancreas segmentation. Patient stratification (healthy and pancreatitis) information is included in the dataset. We share this data with the public, notably the first-ever available pediatric MRI dataset for pancreas research. Please refer to our dataset at NIH supported OSF servers: https://osf.io/km6yz/.

2.2   Segmentation Algorithm (*PanSegNet*)

In this work, we employed the state-of-the-art pancreas segmentation tool, **PanSegNet**, which integrates a novel linear self-attention mechanism into nnUnet architecture to enable efficient volumetric computation (12). The architecture of *PanSegNet* consists of three core components: a standard convolutional encoder, a linear self-attention transformer, and a convolutional decoder (**Figure 2**). This innovative design allows for enhanced feature extraction and improved segmentation performance, particularly in complex medical imaging tasks. For the first time in the literature, *PanSegNet* was trained on a large-scale MRI dataset comprising 767 scans from 499 participants (12). This is the largest dataset used for pancreatic segmentation in MRI-based research to date. The model achieved 85.0% Dice scores on T1W scans and 86.3% on T2W scans, making the highest reported segmentation accuracy in the literature for pancreatic MRI research. These results underscore *PanSegNet*'s capability to perform robust and accurate pancreas segmentation on large, clinically relevant datasets, setting a new benchmark for future studies in the field.

Adhering to the standard nnUNet framework, *PanSegNet* first pre-processes MRI scans through normalization and then automatically configures the network architecture based on the specific dataset characteristics. The training process utilizes stochastic gradient descent (SGD) optimization across five NVIDIA A6000 GPUs, ensuring efficient handling of the large-scale MRI data. The hyperparameters were carefully set, with a learning rate of 0.01, a batch size of 2, and a total of 600 training epochs. Data augmentation techniques were also applied to enhance the model's robustness and prevent overfitting, further improving generalizability (12). In this study, we utilized our publicly available pre-trained *PanSegNet* model for the pediatric population, ensuring accurate and reliable segmentation in pediatric MRI scans. Therefore, we did not perform any internal data splitting within this dataset due to the fact that the algorithm was developed and trained already using an external adult dataset (CT and MRIs) (12), and our presented pediatric cohort in this study was used (only and) exclusively for external testing and validation. This approach ensures no data leakage while comprehensively evaluating model generalizability across anatomical and pathological variations.

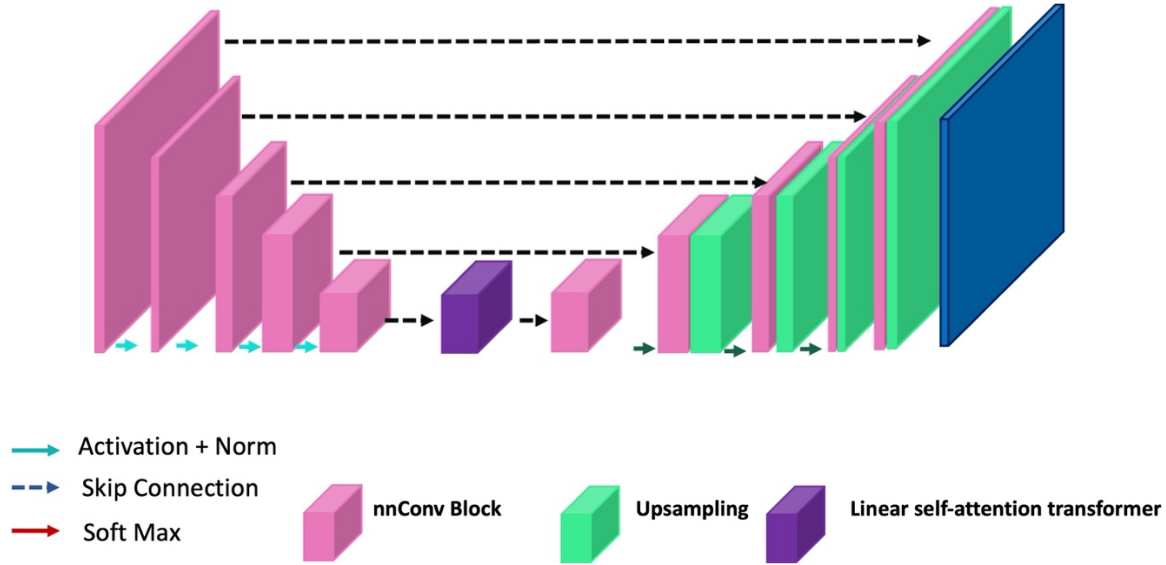

**Figure 2**: *PanSegNet* architecture is based on a combination of nnUnet with a linear self-attention mechanism (35). The architecture accepts volumetric input, appreciating the full pancreas anatomy details compared to pseudo-3D approaches (i.e., slice by slice) (12).

2.3     Software Use

We designed user-interface-based software following the click-and-get rules to provide intuitive support for clinical purposes. Clinical researchers can automatically get the segmentation prediction and volume measurements by selecting the modalities (**Figure 3**).

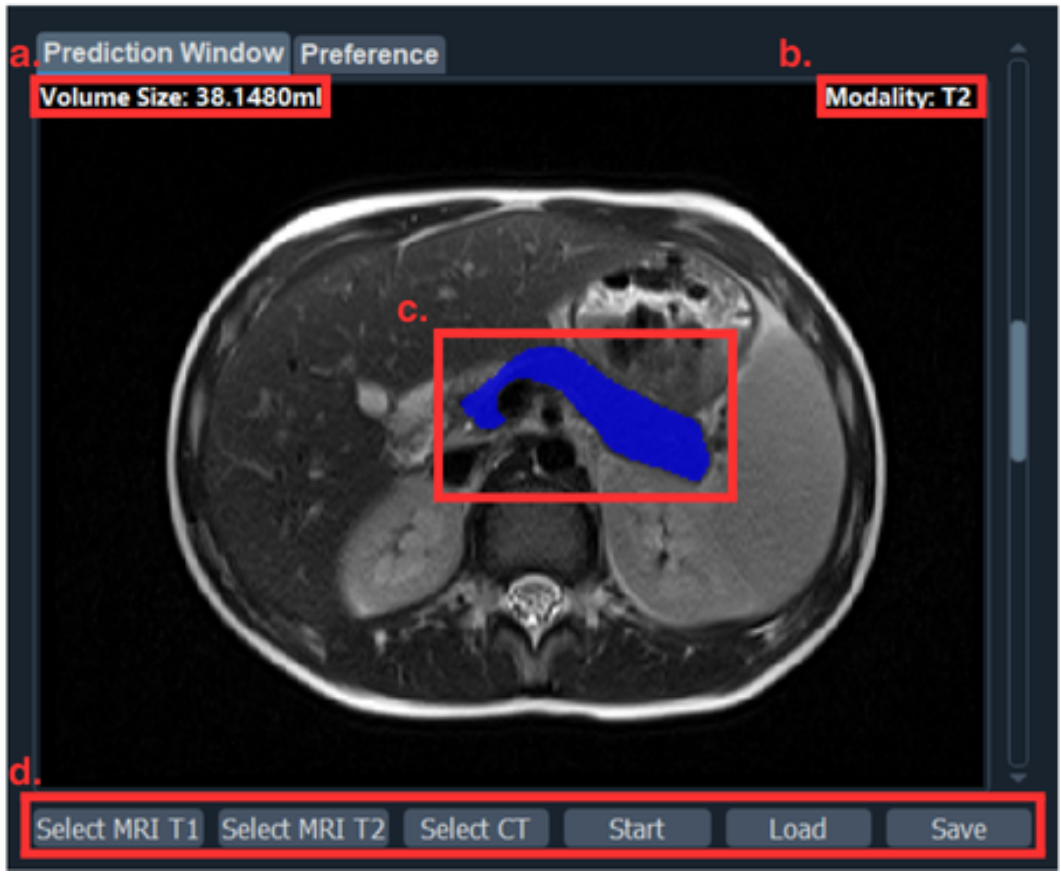

**Figure 3:** User Interface of *PanSegNet* Pancreas Segmentation Software. It reveals *PanSegNet* pancreas segmentation software interface; the volume size indicator (a.) displays the computed volume of the segmented region in milliliters, the modality information (b.) indicates the imaging modality (e.g., MRI T2), the segmented area (c.) represents the region of interest (ROI) automatically segmented by the software, providing precise volumetric analysis, and the control panel (d.) allows for the selection of imaging modalities, initiating segmentation, and managing data (loading and saving).

2.4     Benchmarking with other state-of-the-art segmentation methods

Pancreas segmentation can be approached through multi-organ or single-organ segmentation methods (36, 37). The techniques designed for multi-organ segmentation had suboptimal outcomes for pancreas segmentation compared to those developed for single-organ segmentation approaches. Training a model only on the pancreas facilitates a more specialized procedure, resulting in superior segmentation accuracy compared to a multi-organ model (10). Moreover, single-organ segmentation models may have a simpler architecture than multi-organ models, necessitating fewer training resources and diminishing computing complexity (10, 12). Considering the previous literature, we have focused on a single-organ segmentation model to increase the segmentation accuracy and 3D volume detection.

Other than our presented *PanSegNet*, we adapted and used the following state-of-the-art pancreas segmentation algorithms on our pediatric pancreas MRI dataset: TransUnet (38), nnUNet3D (39), nnUNet2D (39), and SynergyNet (40). All four architectures reached high DSC in pancreas segmentation from CT and MRI scans (12, 18, 41). Among those, nnU-Net is widely recognized and used for its success in various segmentation applications thanks to its enhanced architecture optimization schemes. More recently, hybrid algorithms combining U-Net style algorithms with Transformers gained significant attention as an alternative method to nnUNet because of the self-attention mechanism in transformers shown to improve segmentation results of complex anatomical structures (38). Another state-of-the-art result for pancreas segmentation was achieved by a SynergyNet algorithm, integrating continuous and discrete latent spaces to improve segmentation results (42). We trained all four models on our adult pancreas MRI dataset in the same manner as *PanSegNet* and conducted benchmarking.

**3.RESULTS**

3.1 Segmentation evaluation with DSC and HD95

We evaluated the models' performance on the entire dataset after classifying the data into three groups: healthy subjects, children with AP, and children with CP. In all children, *PanSegNet* showed a strong performance. In healthy children, *PanSegNet* showed more promising results. Similar DSC but lower HD95 were shown in children with AP compared to CP (**Table 2).**

Table 2: *PanSegNet* performance over the entire dataset (84 MRI scans).

|  | DSC (SD) | Precision (SD) | Accuracy (SD) | Jaccard (SD) | Recall (SD) | HD95 (SD) |
|---|---|---|---|---|---|---|
| **ALL** | 0.85 (0.16) | 0.92 (0.12) | 0.99 (0.0) | 0.77 (0.20) | 0.82 (0.19) | 7.94 (17) |
| **Chronic Pancreatitis** | 0.80 (0.20) | 0.90 (0.16) | 0.99 (0.0) | 0.72 (0.23) | 0.80 (0.22) | 15.67 (26) |
| **Acute Pancreatitis** | 0.81 (0.19) | 0.93 (0.11) | 0.99 (0.0) | 0.71 (0.2) | 0.76 (0.23) | 9.85 (19.7) |
| **Normal** | 0.88 (0.1) | 0.92 (0.11) | 0.99 (0.00) | 0.81 (0.16) | 0.87 (0.15) | 3.98 (7.35) |

SynergyNet, nnUNet3D, TransUNet, and nnUNet2D achieved DSC of 0.80, 0.79, 0.78, and 0.70, respectively. While nnUNet3D surpassed *PanSegNet* by a few millimeters in the HD95 metric, overall, *PanSegNet* demonstrated higher DSC values and lower but acceptable HD95 measurements across all datasets. Notably, *PanSegNet* outperformed all benchmarked models regarding the Jaccard Index **(Table 3).**

Table 3: Benchmarking with the other state-of-the-art segmentation methods.

| Model | DSC (SD) | Jaccard (SD) | HD95 (SD) |
|---|---|---|---|
| **PanSegNet** | 0.85 (0.16) | 0.77 (0.20) | 7.94 (17) |
| **nnUNet2d** | 0.70 (0.20) | 0.56 (0.22) | 10.14 (16.11) |
| **nnUNet3d** | 0.79 (0.18) | 0.68 (0.21) | 7.25 (14.24) |
| **TransUnet** | 0.78 (0.12) | 0.69 (0.05) | 8.01 (4.89) |
| **SynergyNet** | 0.80 (1.02) | 0.70 (1.1) | 7.99 (4.32) |

3.2 Pancreas volume assessment

Our volumetric comparisons, which assessed the pancreas volume predicted by the *PanSegNet* algorithm against the corresponding ground truths, revealed significant and robust correlations. *PanSegNet* demonstrated excellent predictive accuracy for healthy subjects, achieving a coefficient of determination ($R^2$) of 0.85, as illustrated by the linear fitting line in **Figure 4A**. In the pancreatitis group, we aimed to ensure the reliability of *PanSegNet*'s volume predictions by excluding poor quality cases (i.e., removing five cases from this analysis). The algorithm achieved an $R^2$ value of **0.77** for patients with acute pancreatitis (AP) and chronic pancreatitis (CP), as shown in **Figure 4B**. **Figure 4C** presents the linear regression across the entire PedPanSegNet dataset, encompassing healthy subjects and those with pancreatitis.

This comprehensive analysis highlights the strong overall performance of *PanSegNet* in predicting pancreas volumes in pediatric MRI scans, regardless of disease status.

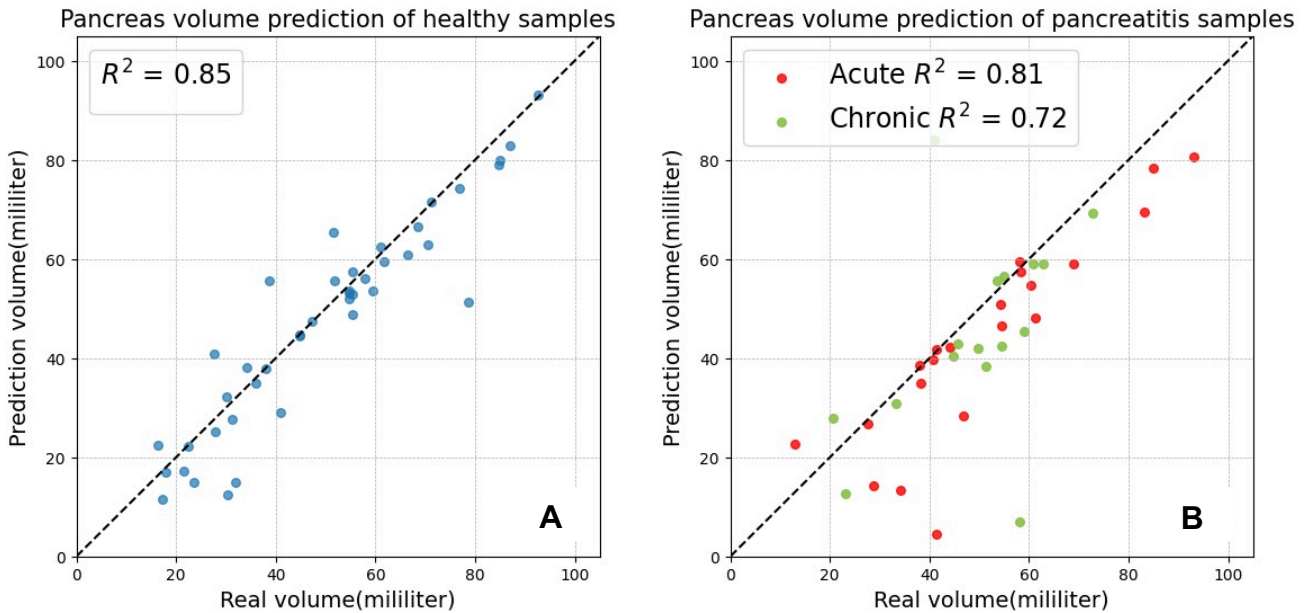

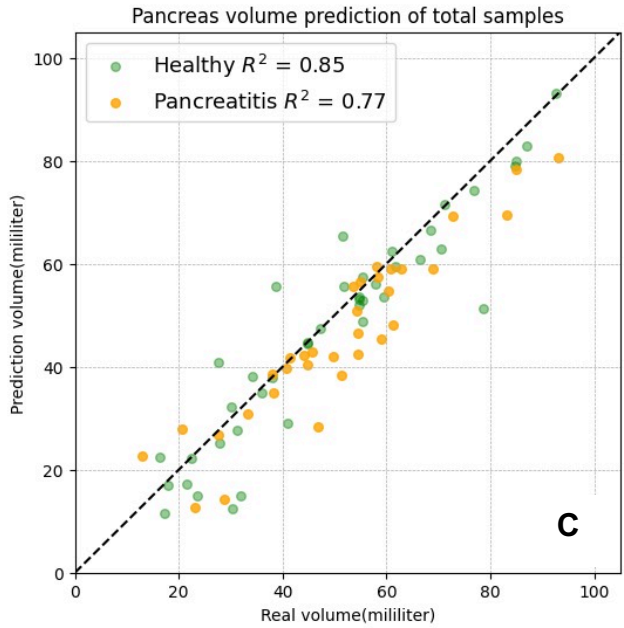

**Figure 4 A, B, C:** Each subplot illustrates a linear fitting line with $R^2$ values of healthy subjects, pancreatitis, and whole PedPanSegNet with real volumes and prediction volumes for the cases.

Our algorithm could not segment the MRI scan of one patient from the AP group and three patients' MRI scans from the CP group. Failure cases (one AP, three CP) are illustrated in **Figure 5.** Consistent with quality-control conventions in radiology, some scans were excluded from the analysis due to low/poor quality (i.e., non-diagnostic). Those cases with low quality were explained on the Supplementary document.

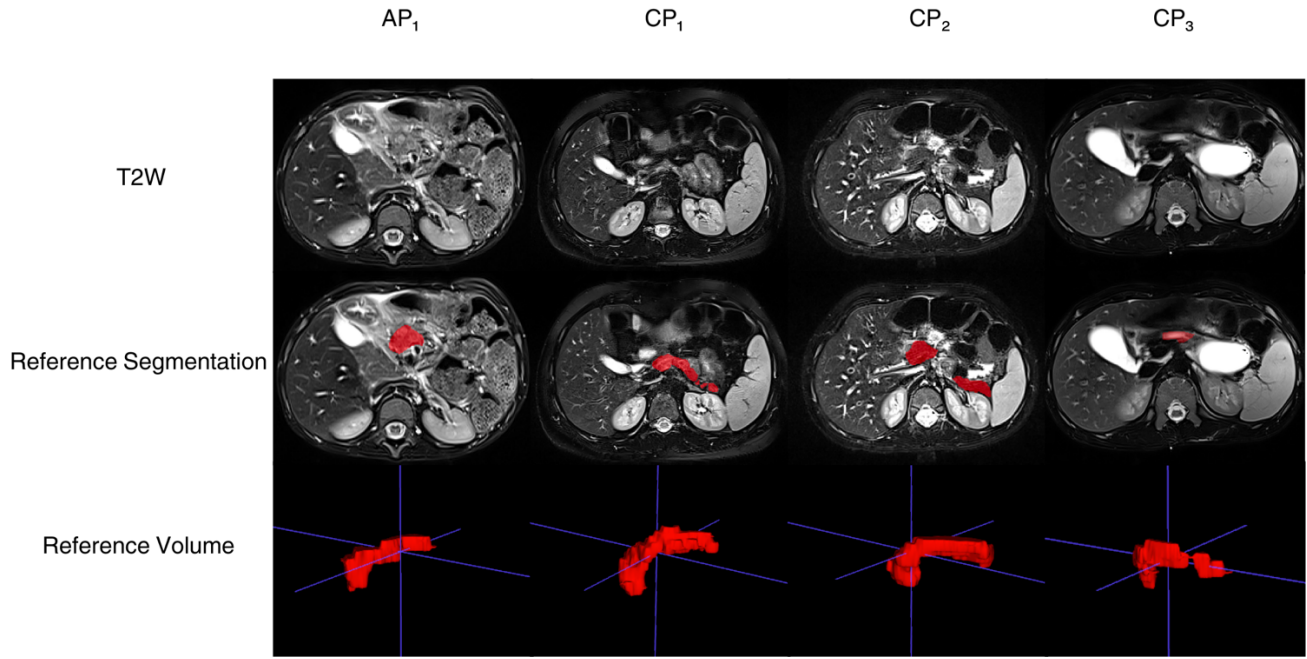

**Figure 5**: Failed pancreas segmentation using *PanSegNet* in acute and chronic pancreatitis patients. Rows: (1) T2-weighted MRI images; (2) segmented T2-weighted MRI images with reference manual segmentation; (3) reference volumes of manual segmentation. Columns: (1) AP case; (2, 3, 4) CP cases.

Along with the precise segmentation provided by *PanSegNet*, our software measured the pancreas volume in milliliters (**Figure 3**). **Figures 4A, 4B,** and **4C** display both the predicted volume and the corresponding ground truth volume.

3.3 Inter and intra-observer agreements

The inter-observer analysis demonstrated a high level of consistency with Cohen's Kappa coefficients of 0.82 and 0.88 on pancreatitis data and normal pediatric data, respectively. Intra-observer analysis of segmentation results was applied to the pediatric pancreatitis data and normal pediatric pancreas data with Cohen's Kappa values of 0.81 and 0.85, respectively, reflecting the stability and reliability of the segmentation models **(Table 4).** Overall, the segmentation task presents considerable challenges.

**Table 4**: Inter and intra-observer evaluations in pancreatitis (blue) and healthy controls (orange).

| Evaluations | DSC (SD) | Precision (SD) | Jaccard (SD) | Recall (SD) | HD95 (SD) |
|---|---|---|---|---|---|
| **Inter-observer** | 0.82 (0.17) | 0.82 (0.19) | 0.77 (0.20) | 0.83 (0.20) | 7.89 (16.40) |
| **Intra-observer** | 0.81 (0.20) | 0.93 (0.11) | 0.73 (0.24) | 0.77 (0.24) | 12.07 (23.01) |
| **Inter-observer** | 0.86 (0.13) | 0.83 (0.18) | 0.81 (0.16) | 0.87 (0.16) | 3.94 (6.53) |
| **Intra-observer** | 0.88 (0.11) | 0.93 (0.11) | 0.79 (0.16) | 0.84 (0.14) | 3.26 (5.40) |

### 4. DISCUSSION

**Key findings and clinical value.** Accurate segmentation of the pancreas in pediatric patients is crucial for diagnosing and managing pancreatic diseases. This study evaluated *PanSegNet*, our effective DL approach for pediatric pancreas segmentation, using T2W MRI scans from patients with AP, CP, and healthy subjects. *PanSegNet* achieved a DSC of 0.88 in healthy subjects and 0.85 in the pancreatitis group, the highest reported DSCs in 3D MRI for pancreas segmentation in a pediatric population to date. In addition to the high-performance segmentation results, our study presents a user-friendly and well-performing DL-based multi-modal pancreas segmentation interface. This interface will be invaluable for clinicians and researchers in pediatric pancreas imaging.

**Volume prediction and clinical tracking.** Furthermore, our study presents the most accurate volumetric prediction of the pancreas segmentation using DL techniques in a pediatric population with an $R^2$ of 0.85 in healthy children and an $R^2$ of 0.77 in children with CP and AP. The accurate quantification of pancreatic volume is essential for the long-term follow-up of patients with AP, ARP, CP, and childhood diabetes (5, 33). Normative pancreas volume values in pediatric populations are revealed and stratified by age, BMI, BSA, weight and height (18). Our *PanSegNet* approach will be promising for quantitative volumetric evaluation of childhood pancreatic disorders. Precise volume measurement, support to optimize clinical care, and early identification of children at risk for ARP and CP-related long-term morbidities can be achieved by precisely monitoring pancreatic volume across several phases of pancreatic diseases (5, 29, 33, 34).

**Segmentation Challenges.** Despite *PanSegNet*'s previous success with peripancreatic edema detection in adult CT, pancreatic segmentation in pediatric MRI remains challenging in the presence of fluid collections. Unlike CT, where edema presents as low-attenuation areas, MRI signal characteristics (especially T2-weighted) may resemble normal soft tissues, increasing false-positive segmentation risk. This is compounded by the smaller, more morphologically variable pediatric pancreas (29). There challenges are clinically relevant, as incomplete segmentation could underestimate disease extent. Specifically, for AP: Segmentation challenges in AP cases arise from diffuse edema, glandular swelling, and peripancreatic fluid collections, often resulting in lower recall due to under-segmentation and blurred tissue boundaries. For CP: Segmentation challenges are associated with fibrotic atrophy, calcifications, and ductal irregularities, leading to higher recall but lower precision as the algorithm may over-segment surrounding fibrotic tissue. The relatively high HD95 in CP reflects substantial boundary mismatch due to these morphological changes. To reflect such challenges in the segmentation algorithm, hence, future improvements may incorporate fluid-aware data augmentation, multi-label segmentation architectures, or contextual attention mechanisms to better distinguish pancreatic tissue from adjacent fluid collections.

**Diagnostic Interpretation and MRI protocol.** While this study focuses on anatomical segmentation, *PanSegNet* enables downstream clinical applications, particularly for CP assessment. These include assisting radiologists in morphologic evaluation of ductal dilatation, glandular atrophy, and calcification patterns. Segmentation framework naturally supports radiomic analysis or diagnostic classification frameworks to enhance clinical decision support too. AP, ARP, and CP lack a standardized MRI protocol, a comprehensive radiological reporting system, and a unified grading framework. Recent consensus reports and recommendations have aimed to address these gaps (5, 29, 30). Furthermore, no standardized timeframe has been established for imaging acquisition in cases of AP, ARP, and CP in the pediatric population (29). This lack of standardization may contribute to discrepancies in interpretations, potentially stemming from the non-uniform application of imaging acquisition timelines. These deficiencies highlight the need for standardized assessment objective radiological grading tools in pediatric clinical evaluations (29).

**MRI sequence (fat suppressed/non-fat suppressed) and field strength.** Patients in our dataset underwent MR imaging using either 1.5T or 3T MRI scanners. Although both modalities adequately visualize the pancreas and pancreaticobiliary ducts, 3T MRI inherently offers a greater signal, theoretically enhancing the visibility of small ducts. However, certain artifacts, notably the dielectric effect, are generally more prominent at 3T (29). Some patients had fat-suppressed sequences, while others had nonfat -suppressed sequences, particularly AP and CP patients had HASTE sequences. Our *PanSegNet* algorithm was trained on mainly non-fat-suppressed images; the lower DSC in the pancreatitis group could be explained by an increased number of fat-suppressed images compared to the healthy dataset. There were no standardized imaging sequences for all patients. In our dataset, fat-suppressed T2-weighted were primarily acquired in children with AP and CP, most of whom were scanned on 1.5 T MRI systems. Fat-suppressed T2-weighted sequences in our cohort often exhibited suboptimal quality due to motion artifacts and inhomogeneous suppression in non-sedated children. Training *PanSegNet* on non-fat-suppressed images improves model robustness and aligns with real-world pediatric imaging practices, particularly in settings where fat suppression is not routinely performed or is technically challenging. Given the **limited availability** of the fat-suppressed sequences in *PanSegNet*, we trained *PanSegNet* in mostly nonfat suppressed adult sequences. This approach aligns with real-world imaging practices, particularly in legacy datasets where fat suppression is not routinely performed.

**Variability in MRI acquisition protocols.** Some patients underwent magnetic resonance cholangiopancreatography (MRCP) with respiratory triggering or gating, and optimal imaging requires regular respirations. Additionally, some imaging studies included only thick slices, while others used thin slices, leading to the lack of standardization in the dataset. Variability in MRI acquisition parameters, including field strength, fat suppression, and slice thickness, significantly impacts model performance. Contrast differences between fat-suppressed and non-fat-suppressed T2-weighted sequences affect pancreatic border visibility, especially in inflamed or fibrotic tissues. In our dataset, fat-suppressed sequences were primarily acquired in pancreatitis patients and often exhibited suboptimal quality due to motion artifacts and inhomogeneous suppression in non-sedated children. This technical variability introduces input heterogeneity that can reduce segmentation consistency. However, one may use harmonization and/or intensity standardization and other correction methods to reduce the heterogeneity effects of MRI scans on segmentation (43-45).

**Segmentation failure analysis.** In our investigation of failed diagnostic cases, we discovered that the T2W images for all patients were fat-suppressed. Notably, one acute pancreatitis patient with a history of intestinal malrotation and prior surgical correction presented with atypical anatomical positioning which the pancreatic head and uncinate process were situated more anteriorly than normal, despite the pancreatic parenchyma appearing homogeneous **(Figure 5, $AP_1$ and Supplementary Figure S1).** This patient also exhibited free intraperitoneal fluid, consistent with the AP diagnosis **(Figure 5, $AP_1$).** These findings suggest that anatomical variations, particularly in patients with congenital anomalies or post-surgical alterations, may lead significantly to segmentation failure. Another chronic pancreatitis patient demonstrated bilateral pleural effusion accompanied by a substantial pseudocyst (60x35 mm) in the pancreatic body and tail, with a TRPV6 gene mutation known to be associated with giant pancreatic pseudocysts with mediastinal extension. Further examination revealed a large cystic lesion occupying the pancreatic body and tail and extending to the spleen, without significant solid components and pancreas is atrophic (**Figure 5, $CP_3$**). The considerable size and extent of this lesion likely contributed to visualization failures during standard imaging assessment. In one particularly challenging case, a patient with an 11-year history of Bartter syndrome and multiple pancreatitis episodes presented with significant pancreatic volume loss **(Figure 5, $CP_1$)**. The presence of respiratory motion artifacts considerably complicated the differentiation between damaged pancreatic tissue and surrounding tissues, hampering accurate assessment and likely contributing to diagnostic inaccuracies. Additionally, a 3-year-old patient's case posed unique challenges due to developmentally limited peripancreatic fat volume **(Figure 5, $CP_2$)**, making the pancreas difficult to distinguish from surrounding structures a characteristic anatomical limitation in pediatric imaging that contributed to diagnostic uncertainty. These factors made it difficult for radiologists to evaluate the images of these three patients **(Figure 5).** In addition to the failed cases, scans from AP patients were excluded due to non-diagnostic image quality, primarily caused by respiratory motion artifacts and overall poor resolution. One of AP patients had additional radiological findings, including perihepatic, peri splenic, and intestinal fluid and pancreatic edema ( Supplementary, Figure S2) and one of them has a type 2 anomaly according to the Komi classification of anomalous union of the pancreaticobiliary duct (AUPBD), with no other associated findings (Supplementary, Figure S7). The diagnosis was established by endoscopic retrograde cholangiopancreatography (ERCP) and subsequently managed with a Roux-en-Y hepaticojejunostomy (Supplementary, Figure S7 ). The third one could not be evaluated due to the motion artifact (Supplementary, Figure S3). Two CP patients were excluded due to low-quality scans: one exhibited extensive edema and a 6×4 cm pseudocyst (Supplementary, Figure S5), while the other demonstrated a non-diagnostic scan (Supplementary, Figures 8a and 8b).

**Robustness of manual segmentation.** In several excluded or poorly performing cases, particularly those with motion artifacts, indistinct margins, or fluid collections, *PanSegNet*'s segmentation challenges aligned with regions requiring consensus review during manual annotation. This suggests the algorithm tends to underperform in cases that are also clinically challenging for human readers. Representative DICOM snapshots of low-quality scans are provided in the Supplementary Material to illustrate typical image quality issues. The complete segmentation failures occurred in cases with severe anatomical distortion that exceeded likely because of the model's training distribution. The failed AP case presented with massive peripancreatic fluid collections and pancreatic necrosis, resulting in loss of normal pancreatic architecture and tissue contrast. The three failed CP cases exhibited severe atrophic changes with extensive fibrosis, calcifications, and ductal irregularities that obscured parenchymal boundaries. Additionally, these cases had suboptimal image quality due to motion artifacts and poor tissue contrast. These failure modes suggest that *PanSegNet*'s current training data may lack sufficient representation of severe pathological variants, highlighting the need for expanded training datasets that include more extreme morphological presentations.

**Dataset size challenges.** The pediatric radiology field and pediatrics are encountering delays in advancing AI technologies, primarily due to a lack of shared datasets, a limited number of pediatric radiologists and pediatric subspecialists, and the necessity for precise data labeling to enhance pediatric datasets (21-26). Although our dataset size is relatively small, it represents the first publicly and fully annotated dataset in the literature. Also, the dataset is derived from a single center in a retrospective manner, does not include racial and ethnic variances, and does not have all laboratory values or symptom start dates of AP, ARP, and CP patients. Our study did not stratify pancreatitis cases by severity or complications, which may influence segmentation performance. Notably, several outlier cases that were excluded involved patients with pleural effusions and significant abdominal fluid collections, which are typically associated with more severe disease. These cases presented atypical anatomy and unclear tissue boundaries, making accurate segmentation particularly challenging. Our analysis focused on whole-organ volumetry as an essential prerequisite for downstream lesion-level analysis. Accurate delineation of ductal dilation, cystic changes, and fatty infiltration requires reliable 3D pancreatic segmentation first. To address all these drawbacks, future studies should incorporate clinical severity grading to better evaluate algorithm robustness across the disease spectrum and include feature-level classification, particularly relevant for CP phenotyping.

**Lack of MRI segmentation of pancreas in pediatric field.** The only pediatric study on pancreas segmentation with DL was done on 27 patients utilizing CT, an imaging modality not commonly used to diagnose CP (17). DSC is reported as 0.73 in the healthy pediatric cases and 0.69 in the pancreatitis cases (17). Current literature extrapolated from adult segmentation studies demonstrated higher DSC in CT modalities than MRI for pancreas segmentation (10, 12). Our study achieved higher performance even though the MRI segmentation is more challenging than CT. In two other pediatric studies using 3D MRI organ segmentation (46), the pancreas was not the primary focus; the dynamics of the lung and thoracoabdominal organs were targeted (47). Another study segmented the kidney from 4D DCE-MRI (46). Neither of the studies shared a dataset or code. Our developed and proposed algorithm, *PanSegNet*, had the highest DSC and the strongest volume correlation in the literature compared to other modalities in adults for pancreas segmentation (12).

**Generalization assessment.** This single-center, single-vendor design may limit generalizability and robustness of the model when applied across heterogeneous imaging protocols or institutions using different MRI scanner vendors. Future studies incorporating multicenter datasets with heterogeneous imaging protocols are essential to evaluate and improve cross-site generalizability. To further advance pediatric pancreas research, we aim to conduct multi-center, multi-national, and prospective cohort studies in future work. Through increased collaboration, future studies could investigate the influence of age, biological sex, race, and ethnicity on pancreas segmentation, volume measurements, and objective radiological interpretation in pediatric pancreatology. These collaborative efforts will facilitate the validation and exploration of innovative approaches to pediatric pancreas imaging, including novel segmentation techniques, transfer learning, foundational models, machine learning algorithms, radiomics methods, and the integration of multi-modal imaging strategies.

**Acknowledgement:**

This work is supported by NIH funding: R01-CA246704, UILCDK127384-SUP, and U01-DK127384-02S1.

**Supplementary material**

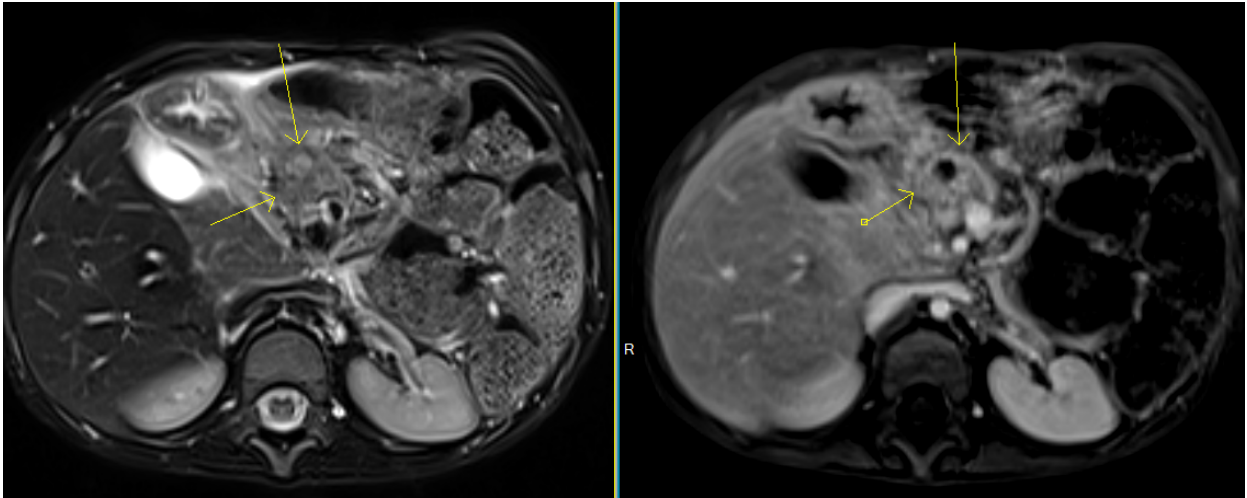

**Figure S1:** *PanSegNet failed to segment the pancreas in an acute pancreatitis ($AP_1$) case, as referenced in Figure 5. Shown are the T2-weighted (left) and T1-weighted (right) MR images.*

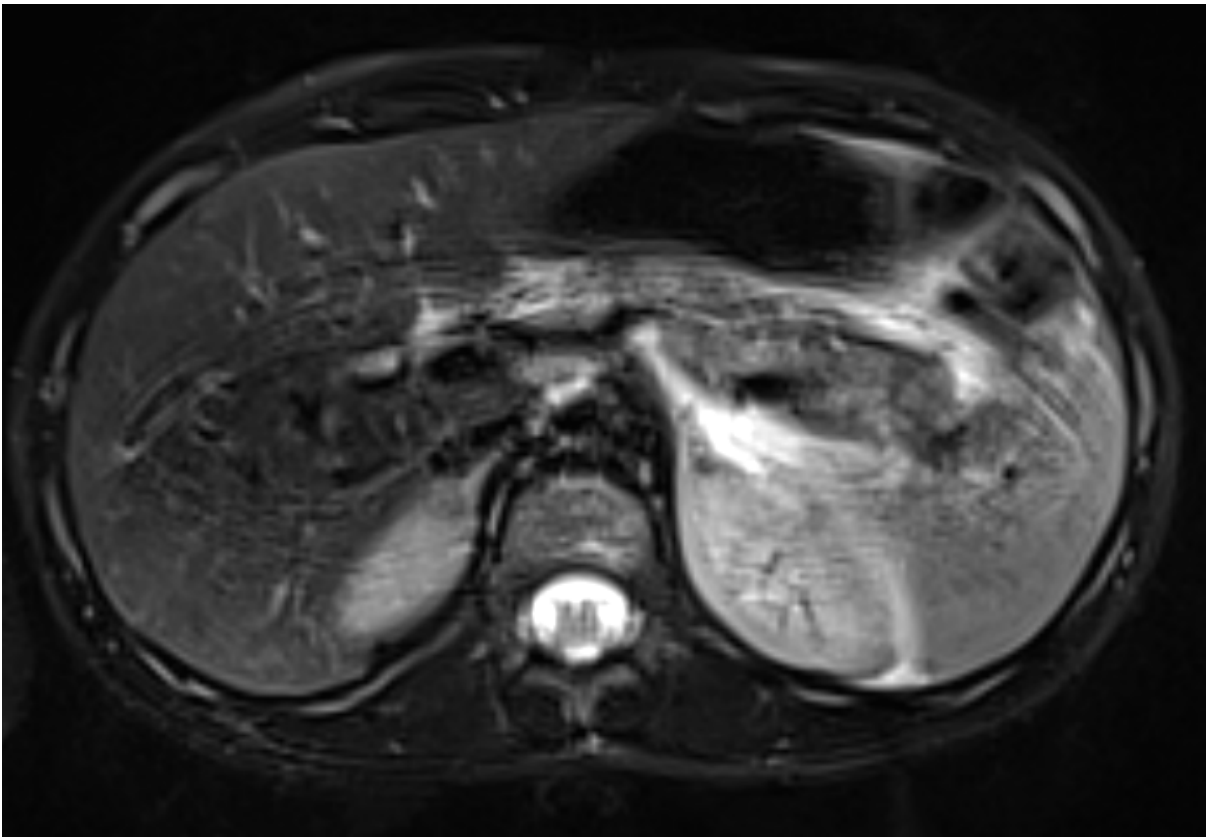

**Figure S2**: Excluded due to the low image quality in AP patient with respiratory motion artifact.

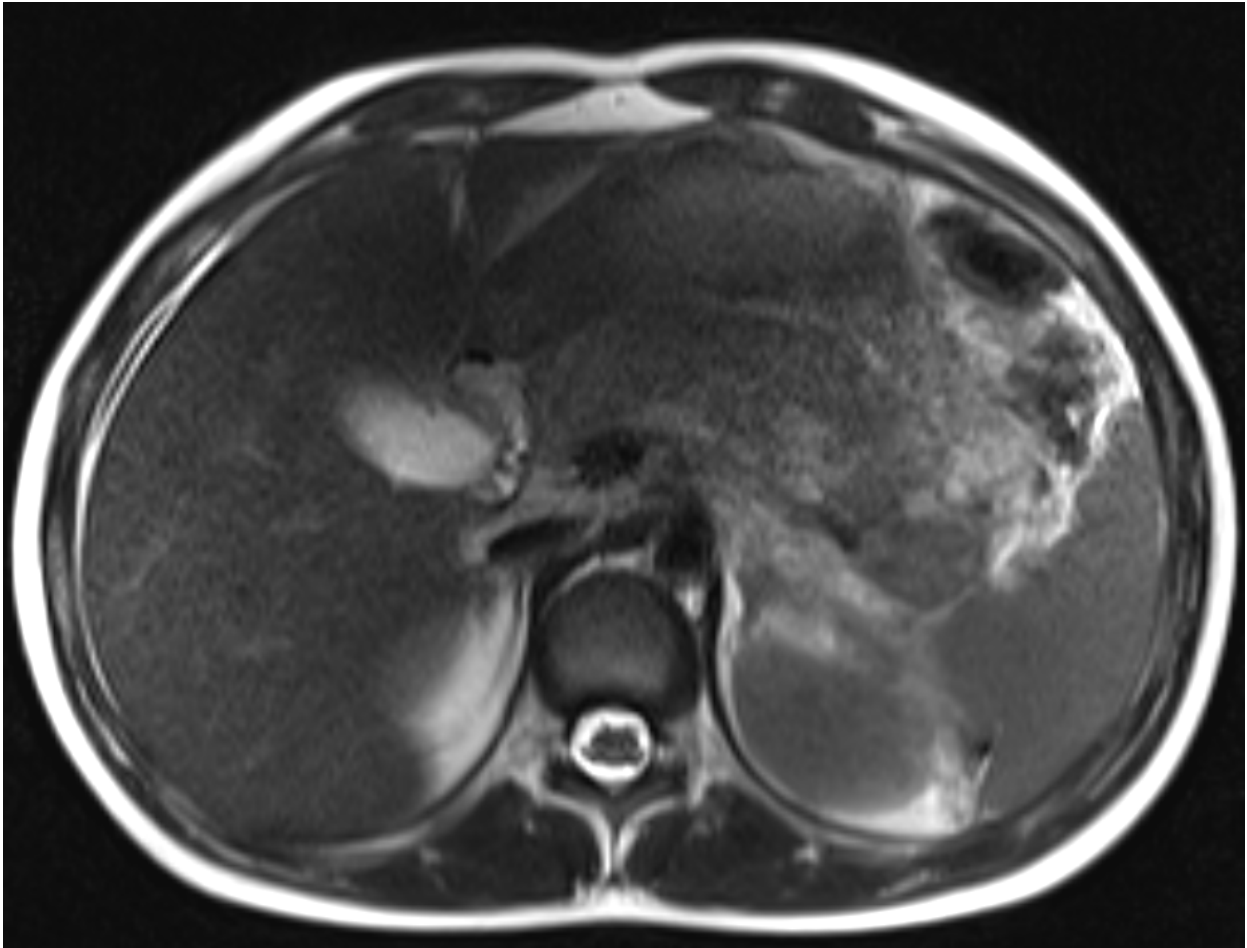
**Figure S3**: Excluded due to the low image quality in an AP patient. The pancreatic body is poorly visualized due to motion artifact related with respiration.

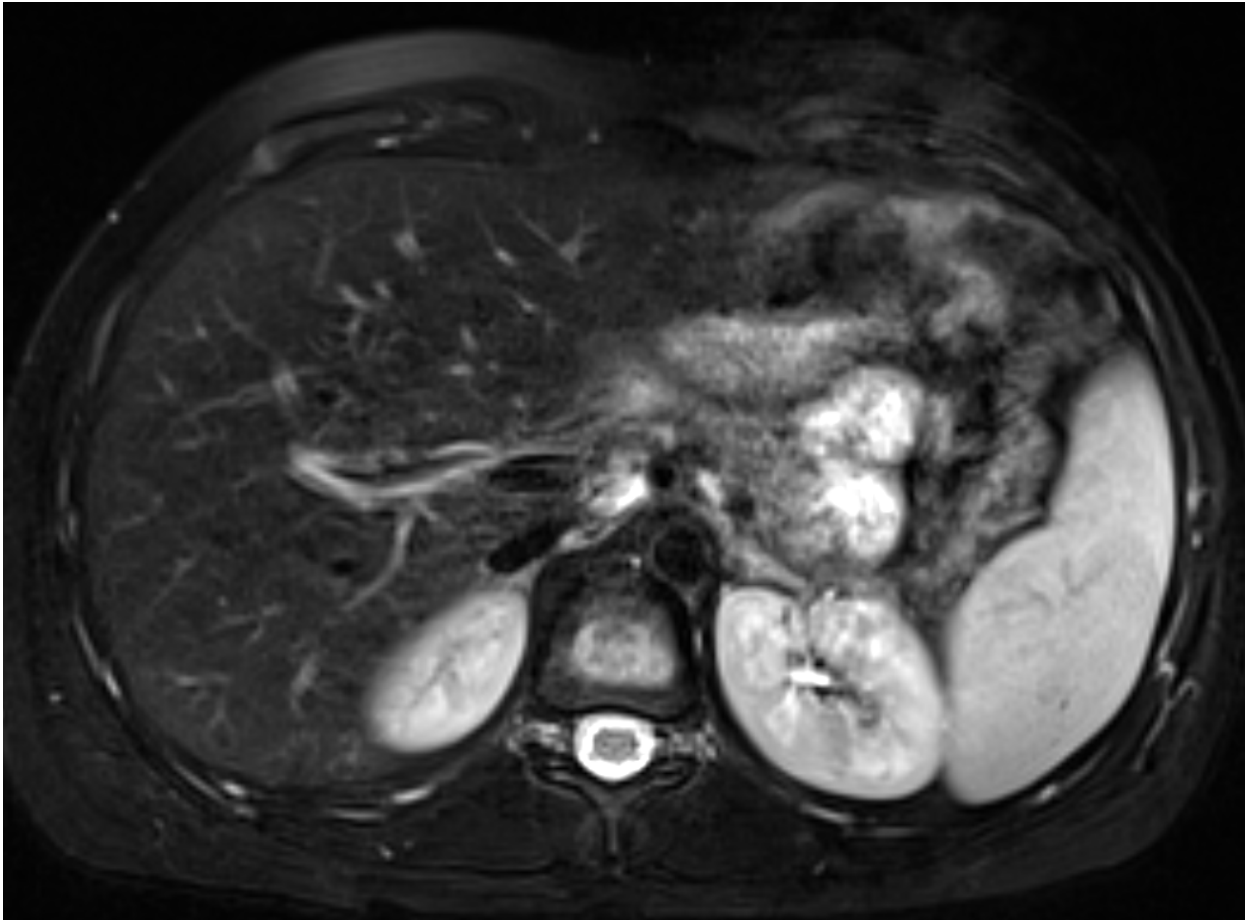

**Figure S4**: *PanSegNet failed to segment the pancreas in a chronic pancreatitis (CP$_1$) case, as referenced in Figure 5.*

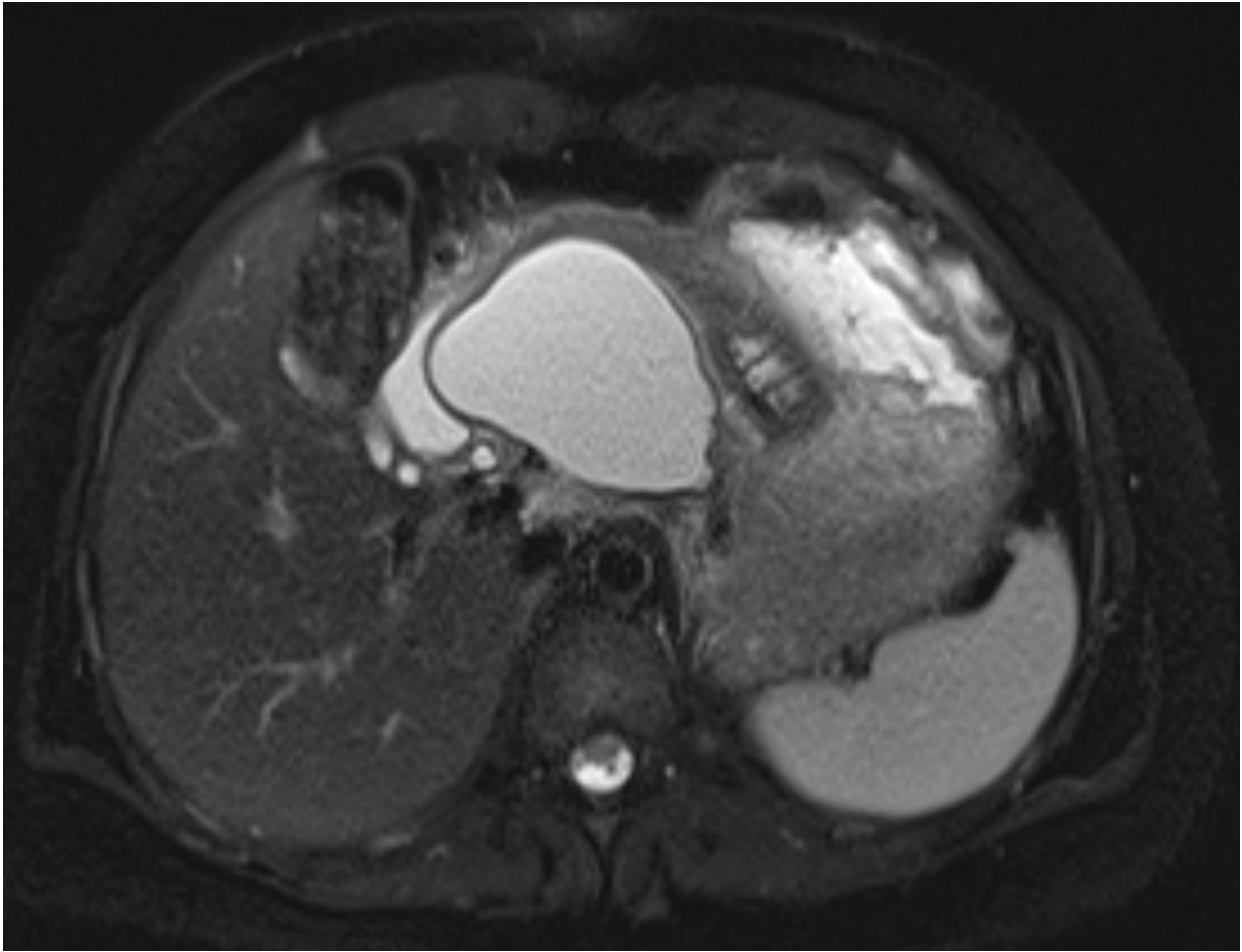
**Figure S5**: Excluded due to low image quality in a CP patient; a 6×4 cm pseudocyst was present in the pancreatic body, obscuring anatomical boundaries.

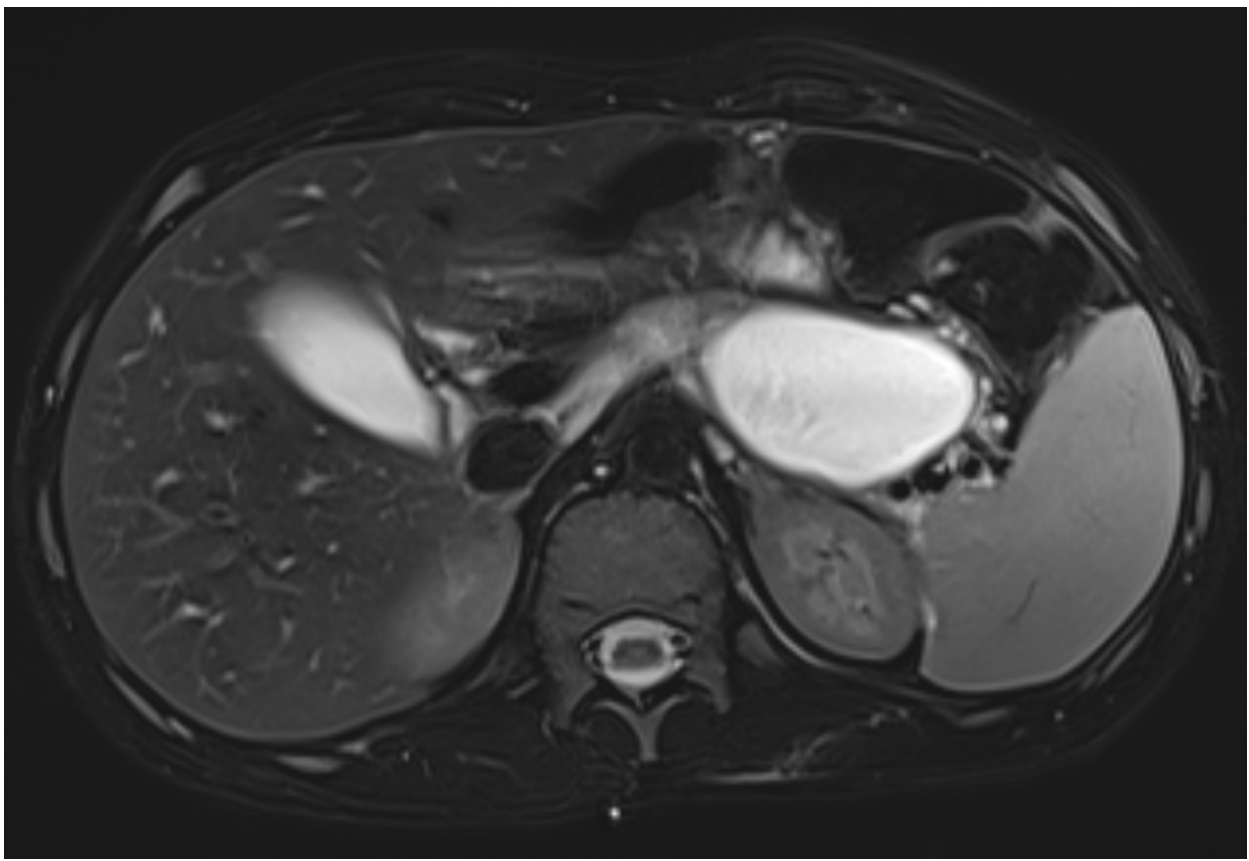

Figure S6: *PanSegNet failed to segment the pancreas in a chronic pancreatitis (CP$_3$) case, as referenced in Figure 5.*

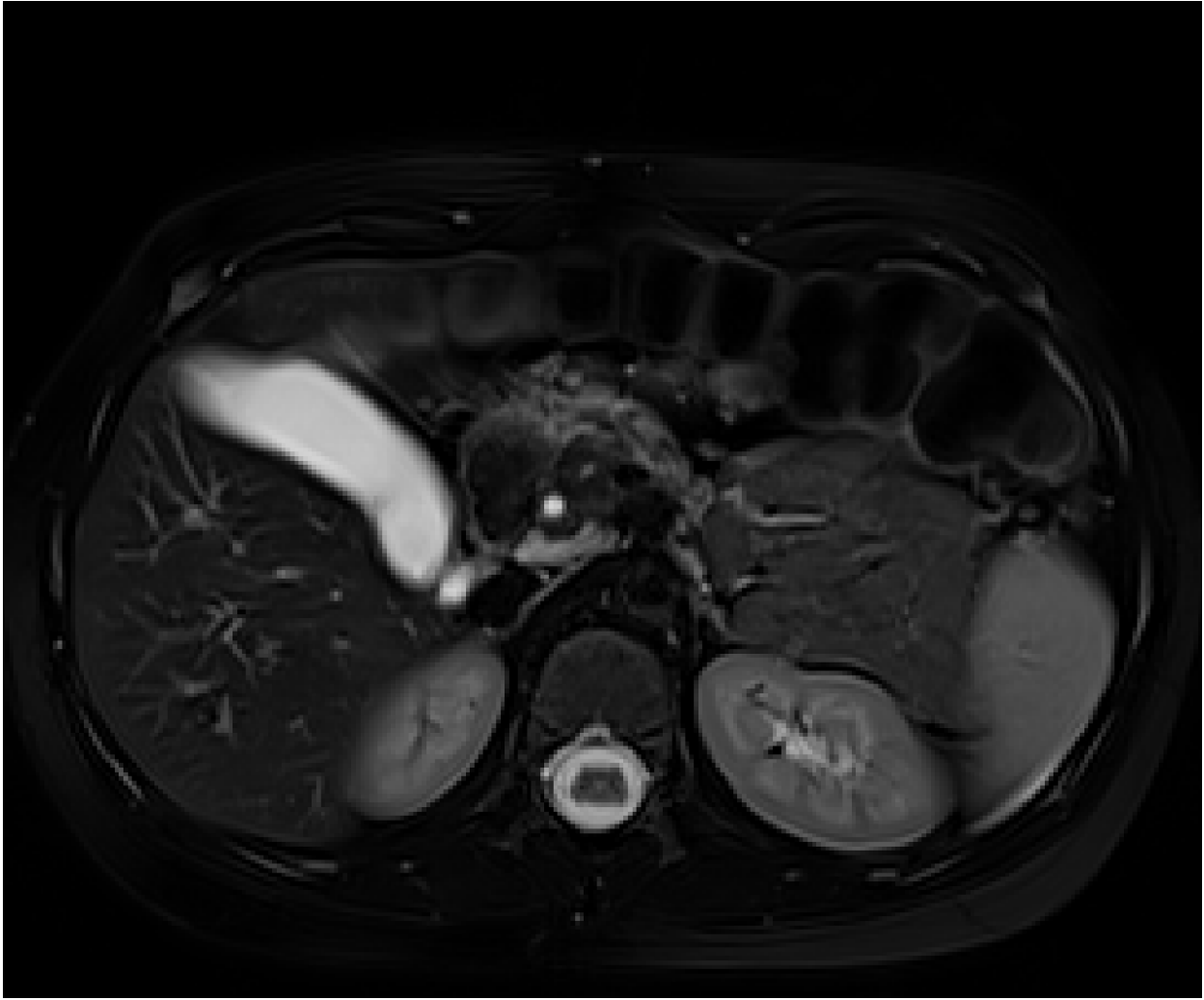

**Figure S7**: Excluded image due to the low quality and non-diagnostic and with type 2 anomaly per Komi classification of AUPBD. No other findings. Diagnosis made by ERCP, followed by Roux-en-Y hepaticojejunostomy.

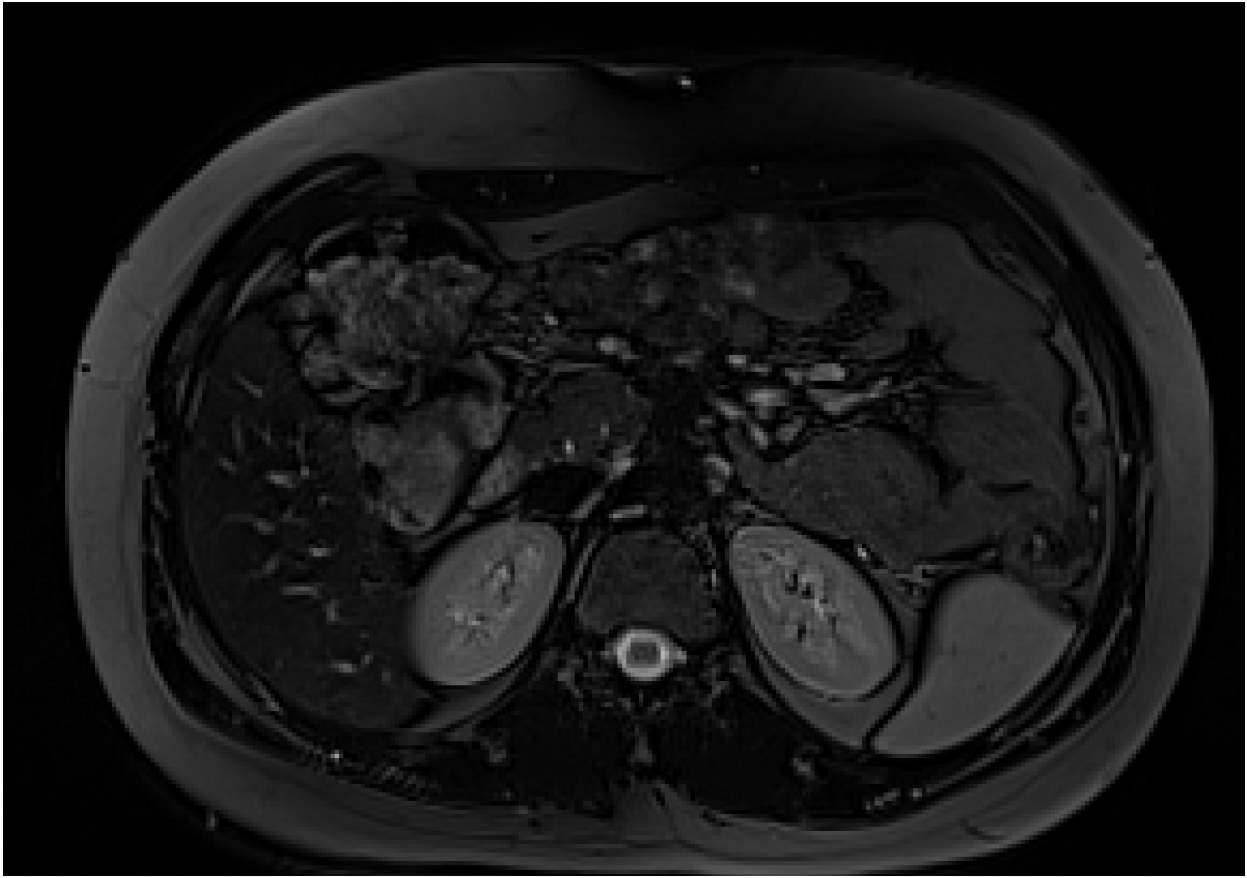

**Figure S8a**: Excluded image due to the low quality and being non-diagnostic, despite a history of chronic pancreatitis, the MRI was reported as normal.

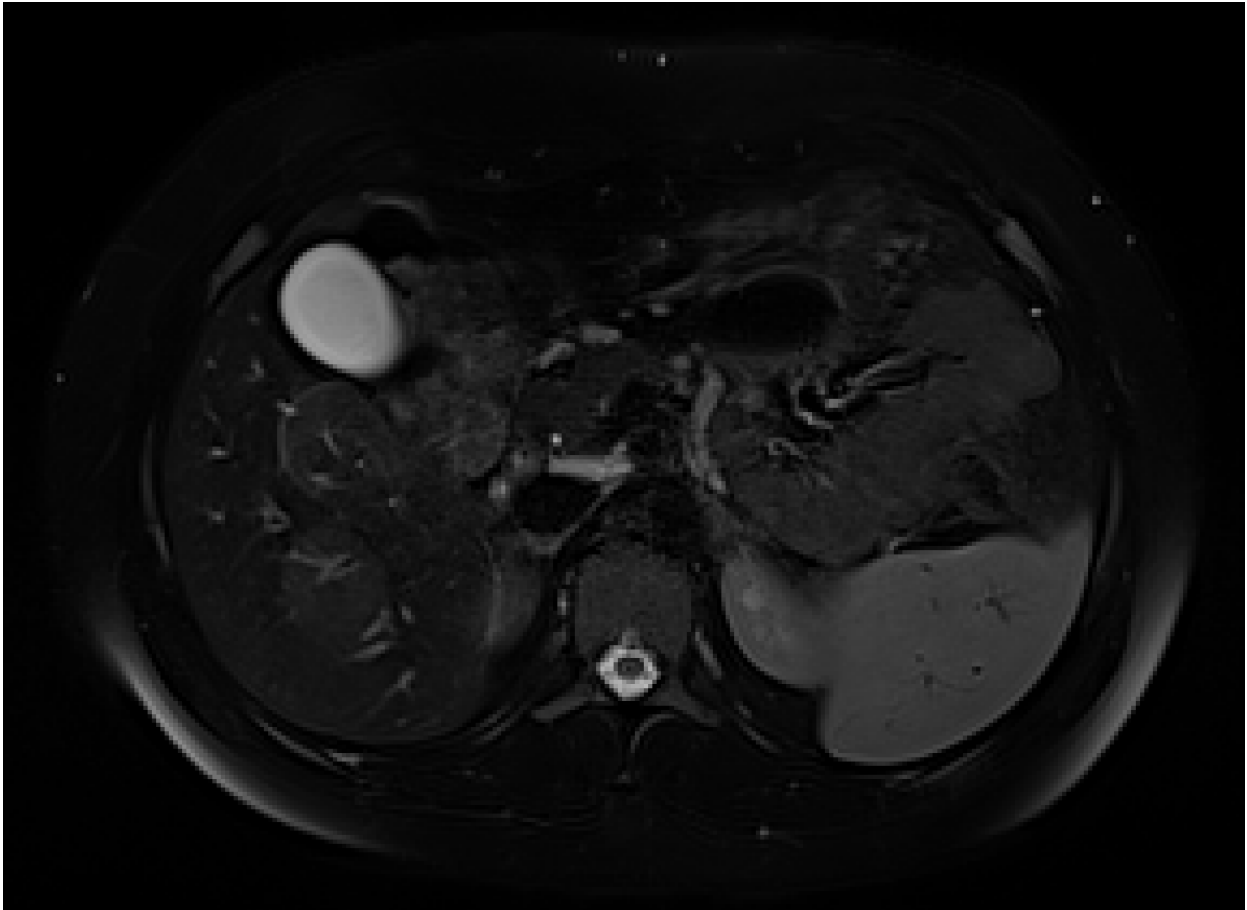
**Figure S8b**: Belonged the subsequent slice to the same patients; not consistent with image acquisition protocol.